\documentclass[10pt,twocolumn,letterpaper]{article}

\usepackage[pagenumbers]{cvpr} 

\usepackage{bbm}
\usepackage{amsmath}
\usepackage{amssymb}
\usepackage{lipsum}

\definecolor{cvprblue}{rgb}{0.21,0.49,0.74}
\usepackage[pagebackref,breaklinks,colorlinks,allcolors=cvprblue]{hyperref}

\def\paperID{18} 
\def\confName{CVPR FGVC12}
\def\confYear{2025}

\title{Predicting butterfly species presence from satellite imagery using soft contrastive regularisation}

\author{Thijs L van der Plas\\
The Alan Turing Institute, the UK\\
Wageningen University \& Research, the NL\\
{\tt\small thijs.vanderplas@wur.nl}
\and 
Stephen Law\\
University College London, the UK\\
{\tt\small stephen.law@ucl.ac.uk}
\and
Michael JO Pocock\\
UK Centre for Ecology \& Hydrology, the UK\\
{\tt\small michael.pocock@ceh.ac.uk}
}

\begin{document}
\maketitle
\begin{abstract}
The growing demand for scalable biodiversity monitoring methods has fuelled interest in remote sensing data, due to its widespread availability and extensive coverage. Traditionally, the application of remote sensing to biodiversity research has focused on mapping and monitoring habitats, but with increasing availability of large-scale citizen-science wildlife observation data, recent methods have started to explore predicting multi-species presence directly from satellite images. This paper presents a new data set for predicting butterfly species presence from satellite data in the United Kingdom. We experimentally optimise a Resnet-based model to predict multi-species presence from 4-band satellite images, and find that this model especially outperforms the mean rate baseline for locations with high species biodiversity. To improve performance, we develop a soft, supervised contrastive regularisation loss that is tailored to probabilistic labels (such as species-presence data), and demonstrate that this improves prediction accuracy. In summary, our new data set and contrastive regularisation method contribute to the open challenge of accurately predicting species biodiversity from remote sensing data, which is key for efficient biodiversity monitoring.
\end{abstract}

\section{Introduction}
There is an increasing demand for resource-efficient methods for monitoring biodiversity at scale. Remote sensing data (such as satellite imagery) promises to meet this demand because of its wide availability and coverage \citep{chandler2017contribution, randin2020monitoring, skidmore2021priority, vanderplas2025ese}, and is routinely used for mapping and monitoring habitats \citep{vanderplas2023multi, garcia2022land}. 
At the same time, its applicability to monitoring animal species biodiversity has largely been confined to large mammals visible from space \citep{attard2024review}, limited by spatial resolution (typically ranging from 10 cm for commercial aerial/satellite imagery to 10 m for public satellite data). 
However, as animal presence depends on the location of species-specific suitable habitats, there is opportunity to \textit{predict} species presence from satellite imagery \citep{he2015will}. Traditionally, this has been done by analysing manually selected satellite-image-derived features, such as land cover \citep{koma2022better, fink2023double} and specific reflectance value indices \citep{koma2022better, weber2018predicting}. But recent work has demonstrated that convolutional neural network (CNN) models are capable of predicting bird \citep{teng2023satbird} and plant \citep{deneu2022very} species presence directly from image data, without the need to manually define the features to be extracted. 
As individual birds or plants cannot be observed directly from 10 m satellite images, ground-truth labels were instead obtained from citizen-science projects where volunteers record and upload geo-referenced records of species they observe \citep{sullivan2009ebird, vanhorn2018inaturalist}. 

In this paper, we curate and publish a new data set of geo-matched satellite images and butterfly observations, evaluate the performance of CNN models, and propose a new soft contrastive learning method that improves predictive performance. 
To create our data set, we used the UK Butterfly Monitoring Scheme (UKBMS) \citep{gbiforg_user_ukbms_2024}, the largest, publicly-available butterfly data set (on GBIF \citep{gbif_general}) of 8M records. 
UKBMS is a high-quality citizen-science data set, which prompted us to ask whether we can improve predictive performance by including a second task that learns from the similarity of species presence between locations. Specifically, we develop a soft, supervised contrastive loss based on the similarity between the probability (species-presence) vector labels, and test whether including this as contrastive regularisation improves learning. To summarise, our contributions are:
\begin{itemize}
    \item We create a new data set called S2-BMS, of geo-matched satellite images and UK butterfly species counts. We have made this data set publicly available as a new benchmark data set for predicting species presence from remote sensing data. 
    \item We propose a novel soft, supervised contrastive loss for contrastive regularisation using the probability vector labels, called Paired Embeddings Contrastive Loss (PECL), which uses the similarity between species-presence vectors to define positive pairs. 
    \item We experimentally optimise a Resnet-based prediction model on the S2-BMS prediction task, and demonstrate that contrastive regularisation improves prediction accuracy. 
\end{itemize}

\section{Related work}
\paragraph{Predicting species presence from satellite data}
Accurate species distribution data is critical for the efficient conservation and restoration of biodiversity \citep{araujo2019standards}. While satellite-data-derived features, such as land cover or spectral indices, are regularly used to predict species distributions \citep{he2015will}, end-to-end predictions from raw satellite data to species distributions remain little explored. Notable developments are GeoLifeCLEF \citep{cole2020geolifeclef, botella2023geolifeclef}, an annually curated data set of plant species and satellite data in Europe, SatBird \citep{teng2023satbird}, a large data set of bird species and satellite data in the USA and Kenya, and SatButterfly \citep{abdelwahed2024predicting}, a data set of butterfly species and satellite data in the USA. Our data set complements these by using species data collected according to a highly standardised protocol (see \cref{sec:data} for more detail), in a new region (the UK), and is only the second such data set that considers butterflies. 
As our data set is most similar to \citep{teng2023satbird, abdelwahed2024predicting}, we use a pre-trained Resnet-18 based model, which was shown to outperform other pre-trained models in \citep{teng2023satbird}.

\paragraph{Contrastive learning}
Contrastive learning has emerged as one of the most successful methods for pre-training convolutional neural networks (CNNs) \citep{oord2018representation, chen2020big, denize2023similarity}. Typically, contrastive learning methods are self-supervised, meaning they create artificial classification tasks by contrasting different augmentations (\ie, transformations such as rotating an image) of the unlabelled input data. For satellite imagery specifically, methods have been developed that contrast different seasons \citep{manas2021seasonal} or different sensors \citep{scheibenreif2022self, prexl2023multi} of the same location, or contrast geographical vicinity \citep{jean2019tile2vec}. Self-supervised contrastive learning is a popular pre-training strategy because annotating data is resource-intensive, thus reducing the amount of labelled data required. 
On the other hand, where large labelled data sets are available, supervised contrastive learning has the potential to outperform self-supervised learning because of its richer reservoir of positive pairs \citep{khosla2020supervised}. In other words, instead of just using augmented (\textit{e.g.}, rotated/blurred/flipped) samples as positive matches, supervised contrastive learning also matches different images of the same class as positive pairs. 
At the same time, self-supervised contrastive learning has developed the soft contrastive loss to overcome the `class collision' problem, \textit{i.e.}, when negative pairs are semantically false negatives \citep{denize2023similarity, feng2022adaptive, prexl2023multi}. Class collision can arise when using hard (one-hot) labels, which cannot represent semantic relations between samples. 

\paragraph{Contrastive regularisation}
Besides using contrastive learning as pre-training task (which typically requires larger data sets than S2-BMS \citep{elnouby2021large}), it can also be used as contrastive regularisation \citep{lee2022contrastive, kim2021selfreg} in a multitask learning setup. The purpose of including the contrastive loss as an auxiliary task, is to encourage a network to learn embeddings that maintain semantic relations while solving the prediction task. Combining tasks in this way can improve performance as features learnt in one task can benefit the other \citep{caruana1997multitask}.
In this paper, we develop a soft, supervised contrastive loss that uses the species-presence probability vectors to determine positive pairs. We implement this as contrastive regularisation alongside the prediction task. 
We further compare our loss function to other existing methods in \cref{sec:meth_comp_soft_cl}.

\section{S2-BMS data set}\label{sec:data}
\begin{figure*}
  \centering
  \includegraphics[width=\linewidth]{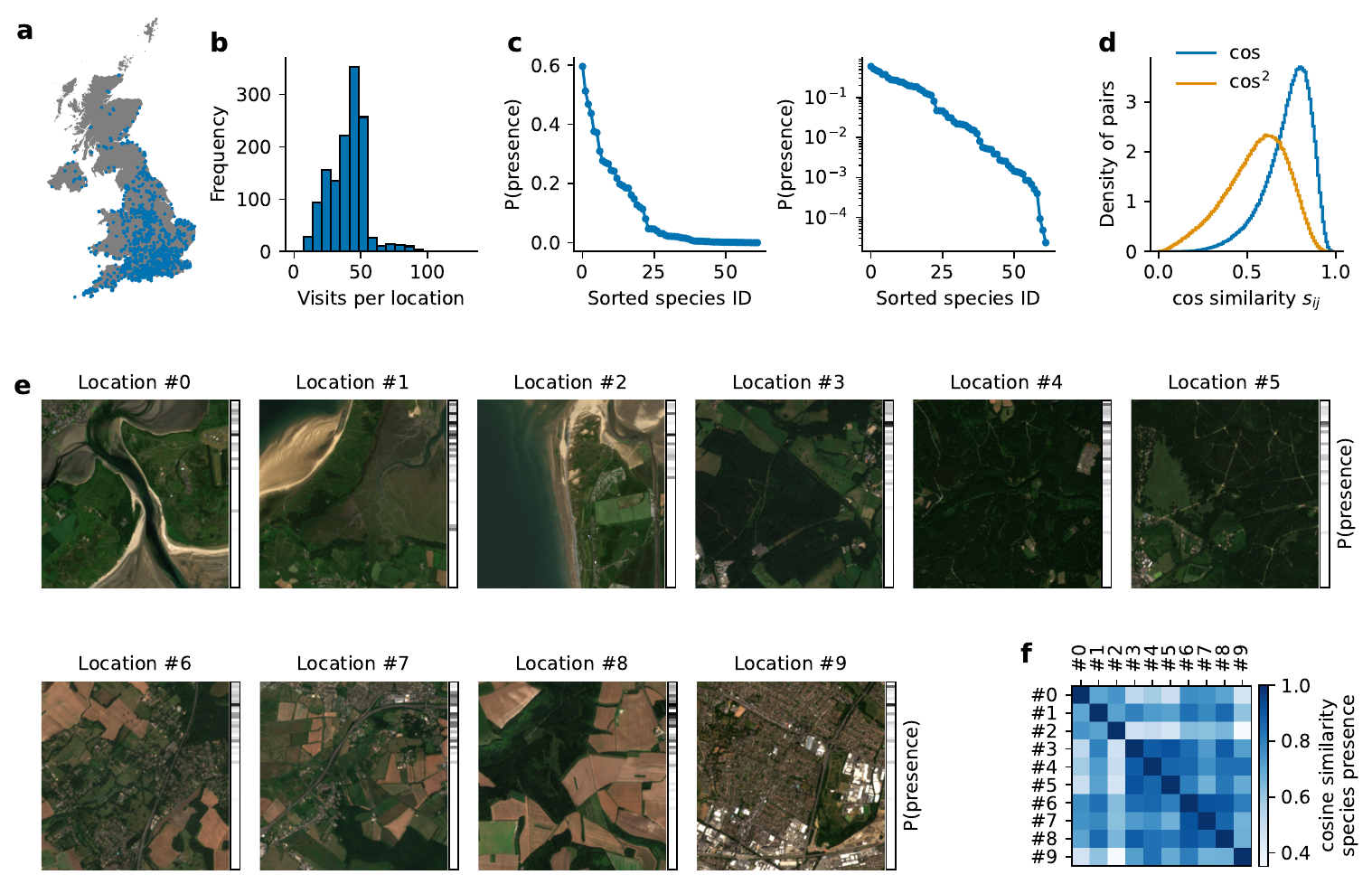}
    \caption{\textbf{Properties of S2-BMS data set.}
    \textbf{a)} Locations (1329 in total) of S2-BMS data set in the UK. 
    \textbf{b)} Number of days that locations were visited over a 2-year period.
    \textbf{c)} (Log) distribution of probability presence per species. 
    \textbf{d)} Distribution of cosine similarity (blue) and cosine similarity squared (orange, used in \cref{eq:loss_softk_pecl}) of species vectors $(\mathbf{y}_i, \mathbf{y}_{j \neq i})$. Total number of pairs is $N (N - 1) / 2 = 882$k.
    \textbf{e)} Example locations of extracted 256 $\times$ 256 pixels S2 images (only RGB shown) and their species-presence vector (right side of each satellite image).
    \textbf{f)} Cosine similarity matrix between the example locations of e).
    }
  \label{fig:dataset_overview}
\end{figure*}

Here we present a new data set called S2-BMS, made publicly available, for predicting species presence from satellite imagery. This data set was compiled from public satellite and biodiversity data, as further detailed below. 

\paragraph{UK Butterfly Monitoring Scheme (BMS)}
The UK Butterfly Monitoring Scheme (UKBMS) is one of the longest running citizen-science projects, monitoring butterflies in the UK since 1976 \citep{UKBMS2018, UKBMS2019}. Data is collected in a very standardised way; the majority of data is obtained by trained volunteers walking the exact same transect route every week (during spring/summer) and logging all butterflies they observe. These data, up to the year 2020, have been made publicly available at GBIF \citep{gbiforg_user_ukbms_2024}.

We used all UKBMS counts of the two most recent, complete years publicly available; 2018 and 2019 (1.12M observations in total). These observations were made at 1804 unique locations (\textit{i.e.}, transects). Each location is stored as a 4 km\textsuperscript{2} square. Locations had 506 $\pm$ 299 observations on average (median $\pm$ $\frac{1}{2}$IQR). We excluded all locations with fewer than 200 observations (to get robust estimates of encounter rates), leaving 1455 locations remaining with 1.08M observations of 62 unique species (\cref{fig:dataset_overview}a). 
For each location, we calculated the fraction of species presence across all visits, in other words, the fraction of visits on which at least one individual (per species) was observed. This yields a vector of encounter rates $\mathbf{y}_i$, of size 62, per location $i$, where each element $y^s_i \in [0, 1]$ is the average P(presence) of species $s$.

\paragraph{Sentinel-2 (S2) satellite data}
We downloaded publicly available Sentinel-2 (S2) reflectance data of the four 10 m resolution bands (Red, Green, Blue (RGB) and Infra-Red (IR)) for the 1455 locations filtered from UKBMS, using Google Earth Engine \citep{sentinel2data}. For each location we downloaded the least cloudy image (using the image property \texttt{CLOUDY\_PIXEL\_PERCENTAGE}) between June 1\textsuperscript{st} and September 1\textsuperscript{st}, for both 2018 and 2019. We visually inspected all images from 2019, and removed those that were still severely obscured by clouds. We then replaced these with cloud-free images from 2018, where available, yielding a final data set of 1329 images. Images are z-scored (per band) upon loading. 

The UKBMS locations are 2 km $\times$ 2 km squares, meaning the transect could have occurred anywhere in that area. To account for this variability, and possibility that the transect may occur at the edge of these squares, we downloaded the S2 images at size 2.56 km $\times$ 2.56 km (= 256 $\times$ 256 pixels), centred at the original squares (see \cref{fig:dataset_overview}e for example squares). During training, we use data augmentation that randomly crops these images to 224 $\times$ 224 pixels. During validation and testing, images are always cropped to the centre 224 $\times$ 224 square. 

\paragraph{S2-BMS data set properties}
Locations are visited typically once every week during the spring/summer season (\cref{fig:dataset_overview}b) by the same trained observer, yielding highly accurate measurements of butterfly presence \citep{UKBMS2018}. Species presence is distributed strongly non-uniformly, \textit{i.e.}, as a long-tailed distribution (\cref{fig:dataset_overview}c), as is typical for species-presence data \citep{teng2023satbird}. Locations are spread throughout the UK, though predominantly in (Southern) England (\cref{fig:dataset_overview}a). \\

Predicting multi-species presence from satellite images is a complex task. Similar habitats often harbour similar species, but differences in habitat condition, geographic location, surroundings etc. can all influence species similarity between locations. To illustrate this, \cref{fig:dataset_overview}f shows the cosine similarity of the species presence labels between the 10 example locations of \cref{fig:dataset_overview}e (and the distribution of all pairs is shown in \cref{fig:dataset_overview}d). 

Our S2-BMS data set differs strongly from, and therefore complements, the SatBird  \citep{teng2023satbird} and SatButterfly \citep{abdelwahed2024predicting} data sets. First, from an ecological perspective, these two data sets consider different geographic regions (the USA + Kenya vs the UK), habitats and taxonomies and/or species. Second, from a data quality perspective, SatBird/SatButterfly have many more locations (123K for the USA in SatBird), but with likely a larger variance in number of visits per location (because UKBMS uses standardised protocols). Further, anyone can log observations on eBird \citep{sullivan2009ebird} and eButterfly (the data providers underlying SatBird and SatButterfly, respectively), while UKBMS collects data via trained volunteers and standardised protocols \citep{UKBMS2018}. Hence, these two data sets differ substantially on a quality versus quantity spectrum.

\paragraph{Train/validation/test split}
We split the data set into 70\% train, 15\% validation and 15\% test splits. To prevent overfitting that could arise from having overlapping/very close locations across splits, we followed the procedure from \citep{teng2023satbird} to spatially cluster locations using DBSCAN \citep{ester1996dbscan} before splitting. All S2-BMS locations whose centre points where less than 4km apart were clustered together, leading to 743 locations grouped into 230 clusters (and 586 locations not clustered). We then randomly split the set of clusters and single locations to create train/validation/test splits of 947/196/186 locations respectively.

\section{Methodology}
The species prediction model is trained to optimise two objectives: a binary cross-entropy loss for predicting species presence and a novel contrastive regularisation loss.
\subsection{Prediction model}\label{sec:meth_pred}
\begin{figure}[t]
    \centering
    \includegraphics[width=\linewidth]{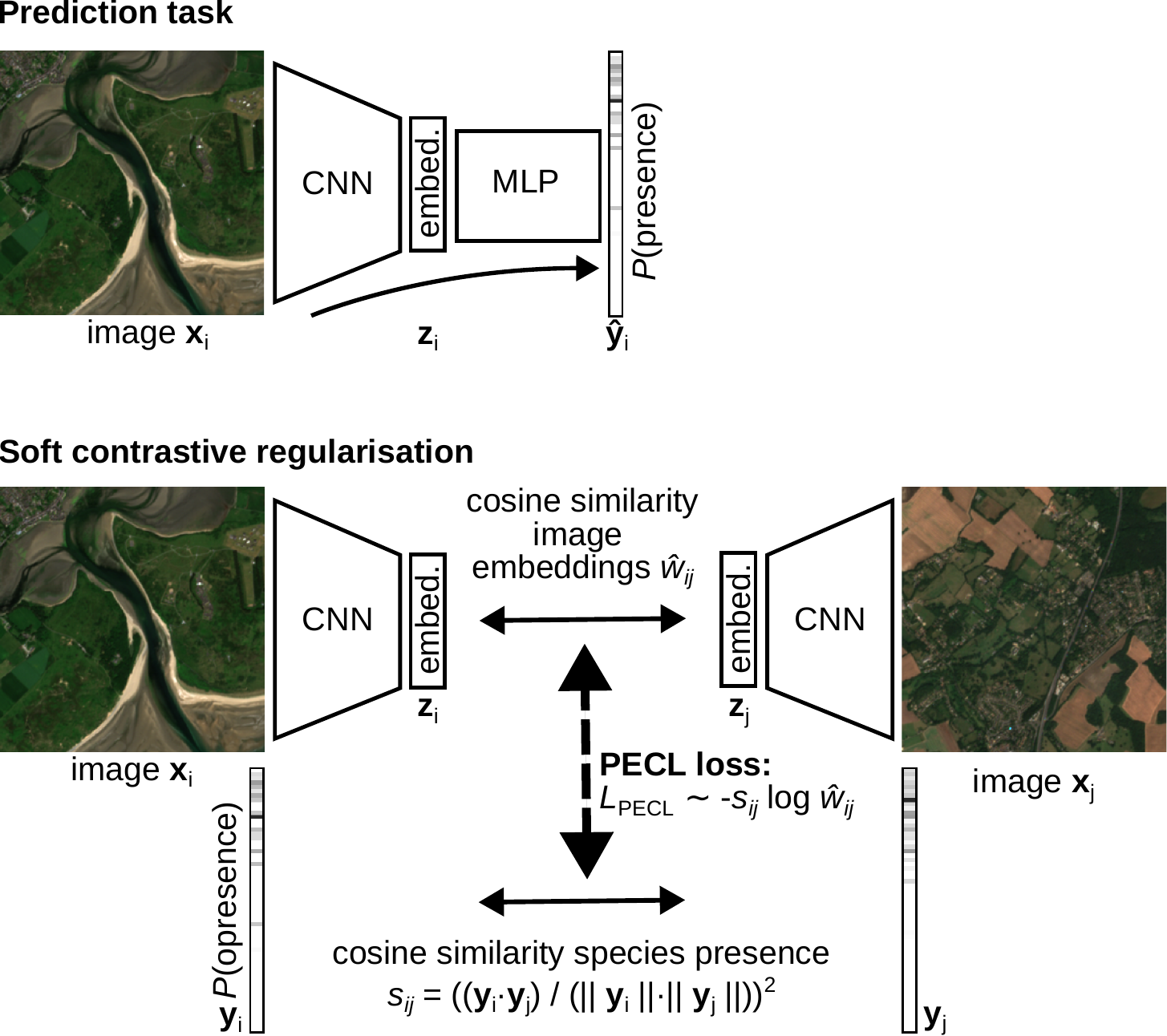}
    \caption{\textbf{Model schematic}. Top: the prediction task (\cref{eq:loss_bce_pred}). Bottom: PECL loss function (\cref{eq:loss_softk_pecl}).}
    \label{fig:schematic_model}
\end{figure}

For an image $\mathbf{x}_i$, we consider its feature embedding $\mathbf{z}_i = f(\mathbf{x}_i)$ of length 256, where $f$ is the (Resnet-18 \citep{he2016resnet} -- the best-performing model in \citep{teng2023satbird}) pre-trained encoder network, and predictions $\hat{\mathbf{y}}_i = g(\mathbf{z}_i)$ of length 62 (= number of species), where $g$ is the multi-layer perceptron (MLP) projector network (\cref{fig:schematic_model}, top), and each $\mathbf{z}_i$ is $l2$-normalised.
To train this prediction model we used a binary cross-entropy (BCE) loss function:
\begin{equation}\label{eq:loss_bce_pred}
    \mathcal{L}_{\textnormal{BCE}} = - \frac{1}{N \cdot S} \sum_{i=1}^N \sum_{s=1}^S \Big( y_i^s \log \hat{y}_i^s + (1 - y_i^s) \log (1 - \hat{y}_i^s) \Big)
\end{equation}
where $y_i^s$ is the true occurrence probability, and $\hat{y}_i^s$ the predicted occurrence probability, of species $s$ at site $i$. $N$ is the number of samples (= sites) and $S=62$ the number of species.

\subsection{Contrastive regularisation}\label{sec:meth_cl}
\paragraph{Supervised contrastive learning}
For conventional contrastive learning (CL), the loss $\mathcal{L}_{\textnormal{CL}}$ is defined using exactly 1 positive sample $p_i$ for sample $i$ \citep{chen2020simple}:
\begin{equation}\label{eq:loss_cl_full}
    \mathcal{L}_{\textnormal{CL}} = - \frac{1}{N} \sum_{i=1}^N \log \frac{\exp (\mathbf{z}_i \cdot \mathbf{z}_{p_i} / \tau )}{\sum_{k \neq i} \exp ( \mathbf{z}_i \cdot \mathbf{z}_k /\tau )}
\end{equation}
where $N$ is the number of samples and $\tau$ is the temperature (for sharpening the distribution). Note that because each $\mathbf{z}_i$ is $l2$-normalised, the inner product $\mathbf{z}_i \cdot \mathbf{z}_j$ is effectively the cosine similarity of $\mathbf{z}_i$ and $\mathbf{z}_j$. \cref{eq:loss_cl_full} can be rewritten as \citep{feng2022adaptive}:
\begin{equation}\label{eq:loss_cl}
    \mathcal{L}_{\textnormal{CL}} = - \frac{1}{N} \sum_i \log \hat{w}_{i p_i}, \quad \hat{w}_{ij} = \sigma(\mathbf{z}_i, \mathbf{z}_j, \tau)
\end{equation}
where $\sigma$ is the soft-max function as defined in \cref{eq:loss_cl_full}. 
In other words, \cref{eq:loss_cl} effectively uses an estimated similarity metric (between two $\mathbf{z}$ samples) $\hat{w} \in [0, 1]$ to compute a cross-entropy loss with one non-zero label $\mathbbm{1}_{j = p_i}$. SupCon \citep{khosla2020supervised} (supervised contrastive learning) can then be expressed as a multi-label extension of \cref{eq:loss_cl} with a set of positive samples $\mathcal{P}_i$:
\begin{equation}\label{eq:loss_supcon}
    \mathcal{L}_{\textnormal{SC}} = - \frac{1}{N} \sum_i \frac{1}{|\mathcal{P}_i|} \sum_{j \in \mathcal{P}_i} \log \hat{w}_{ij}
\end{equation}
Here, $\mathcal{P}_i$ is typically defined as $\mathcal{P}_i = \{ j \mid \mathbf{y}_j = \mathbf{y}_i, j \neq i \}$, \textit{i.e.}, all other samples with the same one-hot class label as sample $i$. 

\paragraph{Paired Embeddings Contrastive Loss (PECL)}
However, species-presence data $\mathbf{y}$ are not one-hot vectors but vectors of encounter rates (with each element between 0 and 1), and therefore we need to define the set of positive pairs differently. 
We define the set of positive pairs as the $k$-nearest neighbours based on cosine similarity ($\text{sim}$) between the labels \citep{feng2022adaptive}, renamed $\mathcal{N}_i^k$ for clarity:
\begin{equation}
\mathcal{N}_i^k = \{ j \mid \text{sim}(\mathbf{y}_i, \mathbf{y}_j) \text{ is among the } k \text{ largest}, j \neq i \}
\end{equation}
Further, we are now able to replace the hard labels $\mathbbm{1}_{j \in \mathcal{P}_i}$ by soft labels, using the \textit{a priori} known similarity between the label embeddings $\text{sim}(\mathbf{y}_i, \mathbf{y}_j) \in [0, 1]$, where $\mathbf{y}_i$ is the species-presence vector at location $i$. We use soft similarity labels $s_{ij} = \left( \text{sim}(\mathbf{y}_i, \mathbf{y}_j) \right)^2$, as these are more gradually distributed than the non-squared $\text{sim}$ (\cref{fig:dataset_overview}d).
We name the resulting loss function Paired Embeddings Contrastive Loss (PECL) (\cref{fig:schematic_model}, bottom). It uses the $k$-nearest neighbours ($k$NN) per sample (in the batch) to determine the positive and negative pairs (all samples inside or outside of the $k$-neighbourhood, respectively), and uses soft ($0 \leq s_{ij} \leq 1$) labels for the positive pairs:
\begin{equation}\label{eq:loss_softk_pecl}
    \mathcal{L}_{\textnormal{PECL}} = - \frac{1}{N} \sum_i \frac{1}{| \mathcal{N}_i^k |} \sum_{j \in \mathcal{N}_i} s_{ij} \log \hat{w}_{ij}
\end{equation}
where $\mathcal{N}_i^k$ is the set of $k$ nearest neighbours of sample $i$.

This loss function is then added to the prediction loss function (\cref{eq:loss_bce_pred}) as contrastive regularisation \citep{kim2021selfreg, lee2022contrastive}:
\begin{equation}\label{eq:loss_cr_alpha}
    \mathcal{L} = \mathcal{L}_{\textnormal{BCE}} + \alpha \cdot \mathcal{L}_{\textnormal{PECL}}
\end{equation}
where $\alpha$ is optimised experimentally. 

\subsection{Comparison to other soft CL methods}\label{sec:meth_comp_soft_cl}
If the similarity labels $s_{ij}$ were all set to 1, PECL would effectively be a special case of SupCon of \cref{eq:loss_supcon} \citep{khosla2020supervised}, where the set of positive pairs is defined by the $k$NN of each sample $\mathbf{y}_i$. Using the similarity labels $s_{ij} \leq 1$ allows to weigh the loss by the semantic similarity per pair. \\

\cref{eq:loss_softk_pecl} is the soft contrastive learning loss function as also derived by \citet[Eq.~5]{denize2023similarity} (but using $\mathcal{P}_i$), who then develop an estimate of $s_{ij}$ based on weighing the learnt similarity $\hat{w}_{ij}$, called similarity contrastive learning (SCE). 
Similarly, soft labels for contrastive learning are also used by adaptive soft contrastive learning (ASCL) \citep{feng2022adaptive}, where -- put briefly -- $s_{ij}$ is approximated using the cosine similarity of the image embeddings of the $k$ nearest neighbours (\textit{i.e.}, $s_{ij} = \text{sim} ( \mathbf{z}_i, \mathbf{z}_j)$), rather than the similarity between the output labels $\mathbf{y}$. Hence, both SCE \citep{denize2023similarity} and ASCL \citep{feng2022adaptive} use unlabelled data $\mathbf{x}$ to estimate the similarity labels $s_{ij}$.\\

Instead of estimating $s_{ij}$, we propose to use a `paired' embedding space of labels $\mathbf{y}$, which is not subject to optimisation, to create the soft labels $s_{ij}$. Here, we leverage the fact that our labels $\mathbf{y}$ are not one-hot vectors -- as is often the case for classification tasks -- but semantic, multi-label vectors of encounter rates where each element can be non-zero.  
Under the assumption that the only task-relevant information in $\mathbf{x}_i$ is the features predictive of $\mathbf{y}_i$, this thus creates a ground-truth similarity metric.  \\

\section{Experimental details}
\subsection{Mean encounter rate model}
As a baseline, we consider a mean encounter rate model \citep{teng2023satbird}. This model calculates the mean P(presence) for each species across locations in the training set, and uses these rates as predictions for the validation and test sets (regardless of location).

\subsection{Prediction model}
We fine-tuned convolutional neural network (CNN) models to predict species presence from satellite images (\cref{fig:schematic_model}, top). Specifically we used Resnet-18 \citep{he2016resnet} models that took $(4, 224, 224)$-sized images $\mathbf{x}$ as input and transformed these to a 256-dimensional feature embeddings $\mathbf{z}$ (where each feature embedding vector $\mathbf{z}_i$ was $l2$-normalised). We used pre-trained models (ImageNet \citep{torchvision2016}, SeCo \citep{manas2021seasonal}), and froze all Resnet layers except the final (fully connected) layer during training, except where specifically noted otherwise. 
Feature embeddings $\mathbf{z}$ were then used to predict the output vector $\hat{\mathbf{y}}$ via a fully connected MLP network with 3 layers (after optimisation in \cref{tab:mlp-layer_pretrained}), 256 hidden units per layer, ReLU activation functions for intermediary layers and a sigmoid activation function for the final output layer. As a result, both the label $\mathbf{y}$ and the predicted species-presence vectors $\hat{\mathbf{y}}$ were vectors with each element between 0 and 1. 

(Z-scored) input images were augmented during training by random horizontal and vertical flipping and random cropping (from size 256 $\times$ 256 to 224 $\times$ 224). Other common augmentation procedures such as blur and contrast/brightness jitter were found to have negligible effects during preliminary experiments and therefore not used. 
We used an Adam optimiser using an initial learning rate of 0.001 and batch size of 64 with corresponding results shown in \cref{tab:mlp-layer_pretrained}. These two hyperparameters were then tuned and optimised (see \cref{sec:CR_results} for optimal values). We used a BCE loss function for the prediction network (\cref{eq:loss_bce_pred}) (weighted BCE was also tried but did not improve prediction results). We trained for 50 epochs or until convergence of the validation loss, and evaluated the best-performing model (based on the validation set) on the test set. 
The random seed was fixed, and (the same) 3 seeds were tested for all simulations. Results are stated as the mean $\pm$ standard error of the mean (SEM) across these seeds.

\subsection{Contrastive regularisation}
At each epoch the training data set was shuffled such that batches differed across epochs, essentially as augmentation procedure to prevent the $k$NN contrastive learning from overfitting on any particular batch composition. 
We performed hyperparameter searches to optimise the model performance, using both grid and random search. For both search methods, we trained the model using the same 3 random seeds as before, and calculated mean $\pm$ SEM across seeds.

\paragraph{Hyperparameter grid search}
We first optimised the contrastive regularisation hyperparameter using a limited grid search: number of nearest neighbours $k \in \{1,2, 5\}$, ratio losses $\alpha \in \{0.1, 0.3\}$, for learning rate 0.001, batch size 32 and temperature $\tau = 0.5$. 

\paragraph{Hyperparameter random search}
We then attempted to further optimise the model by performing random hyperparameter search, which is known to be more efficient for finding optimal hyperparameters (but makes it difficult to systematically compare between hyperparameter settings) \citep{bergstra2012random}. We generated 30 random hyperparameter settings, by sampling the following parameters, where $U(a, b)$ is a uniformly sampled random number between $a$ and $b$ and $\{\dots\}$ denotes a fixed set:
\begin{itemize}
    \item Learning rate $\text{LR} \sim 10^{U(-5, -3)}$.
    \item Batch size $\sim \{ 8, 16, 32, 64 \}$.
    \item Number of nearest neighbours $k \sim \{1, 2, 3, \dots, 9, 10 \} $, unless batch size = 8, in which case the upper limit for $k$ is 7. 
    \item Ratio losses $\alpha \sim 10^{U(-\frac{3}{2}, 0)}$.
    \item Temperature $\tau \sim U(\frac{1}{10}, 1)$.
\end{itemize}

\subsection{Metrics}
We evaluate each model by calculating the mean squared error (MSE), top-5 accuracy and top-10 accuracy of the test set (following \citep{teng2023satbird}). The top-5 accuracy indicates the percentage of species present in both the top-5 of the predicted label $\hat{\mathbf{y}}_i$ and the top-5 of the true label $\mathbf{y}_i$, averaged across locations $i$. (And equivalent for top-10 accuracy).

\section{Results}

\subsection{Prediction model}
We first optimised the prediction model architecture by varying the pre-trained Resnet weights (Imagenet \citep{torchvision2016} or SeCo \citep{manas2021seasonal}) and the number of layers in the MLP (\cref{fig:schematic_model}). We found that differences were small, except for 1-layer MLPs which had a significantly higher MSE (\cref{tab:mlp-layer_pretrained}). We also lowered the learning rate which did not improve performance.
As a baseline we computed a mean rate model, which performed worse than all CNN-based models (\cref{tab:mlp-layer_pretrained}). In subsequent experiments we used the model that minimised the MSE: the 3-layer SeCo model. 

Further, we trained the 3-layer SeCo model without freezing the convolutional layers, but this did not improve performance (\cref{tab:model_changes}). We observed that for these models, after the validation loss converged, they continued to (over)fit on the training data set, possibly caused by the small size of the data set. Again, using a smaller learning rate did not improve performance. Hence we used frozen Resnet models going forward.

\begin{table}[ht]
\footnotesize
\begin{tabular}{lllll}
\toprule
L & Model & Top-10 [\%] & Top-5 [\%] & MSE [1e-02] \\
\midrule
- & Mean rate & 67.3 & 58.7 & 1.39 \\
\midrule
1 & ImageNet & 68.9 $\pm$ 0.8 & 60.9 $\pm$ 1.0 & 1.33 $\pm$ 0.05 \\
1 & SeCo & 69.1 $\pm$ 0.5 & 61.8 $\pm$ 0.9 & 1.34 $\pm$ 0.04 \\
2 & ImageNet & \textbf{69.7 $\pm$ 0.5} & 61.5 $\pm$ 0.7 & 1.24 $\pm$ 0.03 \\
2 & SeCo & 69.5 $\pm$ 0.4 & \textbf{62.4 $\pm$ 0.8} & 1.22 $\pm$ 0.04 \\
3 & ImageNet & 69.0 $\pm$ 0.5 & 61.8 $\pm$ 0.7 & 1.24 $\pm$ 0.03 \\
3 & SeCo & 69.6 $\pm$ 0.2 & 62.4 $\pm$ 1.0 & \textbf{1.21 $\pm$ 0.04} \\
\bottomrule
\end{tabular}
\caption{Mean and SEM of validation metrics for different pretrained networks and varying number of MLP prediction layers L. The best performing model for each metric is highlighted in bold.
\label{tab:mlp-layer_pretrained}}
\end{table}

\subsection{Contrastive regularisation}\label{sec:CR_results}
We then added contrastive regularisation following \cref{eq:loss_softk_pecl} and \cref{eq:loss_cr_alpha}. We performed a grid hyperparameter search of $\alpha$ and $k$, yielding best results for $\alpha=0.1$ and $k=5$ (\cref{tab:cr_32}). However, differences between different hyperparameter sets were generally very small and insignificant. Compared to the model without contrastive regularisation, only top-10 accuracy slightly improved (\cref{tab:model_changes}). 
We then attempted to find the optimal model hyperparameters using random search, which is generally more efficient for finding optimal hyperparameters (but makes it difficult to systematically compare between hyperparameter settings) \citep{bergstra2012random}. The best random search hyperparameter combination indeed outperformed the best grid search model, and hence was the best model overall (\cref{tab:model_changes}), with learning rate 0.00029, batch size 64, number of nearest neighbours $k = 2$, ratio losses $\alpha =0.06$, and temperature $\tau = 0.72$.

\begin{table}[ht]
\footnotesize
\begin{tabular}{rllll}
\toprule
$k$ & $\alpha$ & Top-10 [\%] & Top-5 [\%] & MSE [1e-02] \\
\midrule
- & 0 & 69.6 $\pm$ 0.2 & 62.4 $\pm$ 1.0 & 1.21 $\pm$ 0.04 \\
1 & 0.1 & 69.8 $\pm$ 0.7 & 62.4 $\pm$ 0.8 & \textbf{1.21 $\pm$ 0.03} \\
2 & 0.1 & 69.6 $\pm$ 0.5 & 62.4 $\pm$ 0.6 & 1.23 $\pm$ 0.02 \\
5 & 0.1 & \textbf{70.0 $\pm$ 0.2} & 62.3 $\pm$ 0.6 & 1.21 $\pm$ 0.03 \\
1 & 0.3 & 69.7 $\pm$ 0.6 & \textbf{62.5 $\pm$ 0.6} & 1.21 $\pm$ 0.04 \\
2 & 0.3 & 69.9 $\pm$ 0.8 & 62.3 $\pm$ 0.7 & 1.23 $\pm$ 0.04 \\
5 & 0.3 & 69.6 $\pm$ 0.6 & 62.1 $\pm$ 0.6 & 1.23 $\pm$ 0.03 \\
\bottomrule
\end{tabular}
\caption{Mean and SEM of validation metrics for networks with and without contrastive regularisation, for various hyperparameter settings. Temperature $	\tau = 0.5$ was used, and batch size 32 for all $\alpha > 0$ contrastive regularisation models.
\label{tab:cr_32}}
\end{table}

\subsection{Results split per location and species}
We assessed how model performance broke down across species and locations by computing $f_{\text{MSE}} = \frac{\text{MSE mean rate}}{\text{MSE model}}$ per location and per species. In other words, $f_{\text{MSE}}$ is the factor that the MSE loss improves by the CNN model compared to the mean rate model baseline, which we calculated per location and per species. We used the best model configuration from \cref{tab:model_changes} and calculated the mean $f_{\text{MSE}}$ values across the random seeds. 

\begin{figure}[b]
    \centering
    \includegraphics[width=\linewidth]{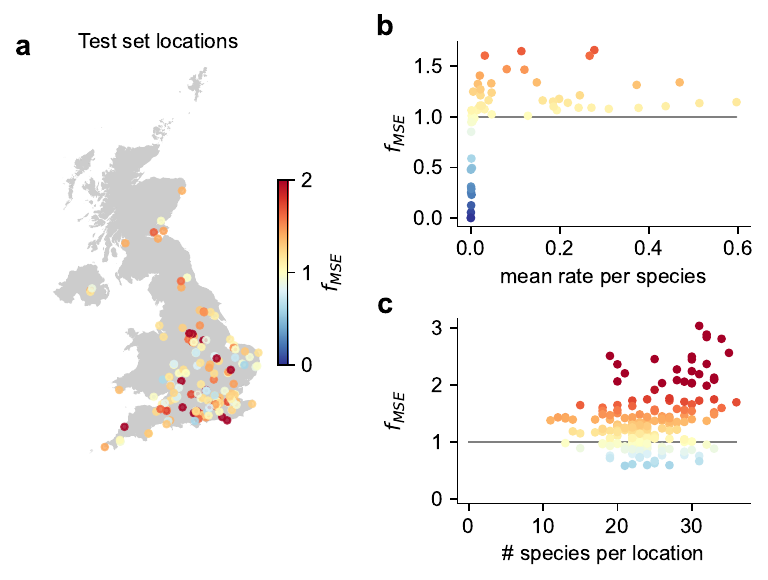}
    \caption{\textbf{Performance across locations and species}, quantified using the MSE improvement factor ($f_{\text{MSE}}$ across locations and species. \textbf{a)} Map of $f_{\text{MSE}}$ values for each test location (values greater than 2 shown as 2), across the UK. \textbf{b)} $f_{\text{MSE}}$ as a function of mean encounter rate per species. \textbf{c)} $f_{\text{MSE}}$ as a function of number of species present per location (\textit{i.e.}, number of species ever observed per location).}
    \label{fig:mse_improvements}
\end{figure}

We found that high-$f_{\text{MSE}}$ locations were distributed across the UK, indicating that good predictions were not due to overfitting to one particular geographic area (\cref{fig:mse_improvements}a). All species with $f_{\text{MSE}} < 1$ had mean encounter rates smaller than 0.01 (\cref{fig:mse_improvements}b), probably due both the lack of sufficient data points as well as the strong class imbalance. Finally, $f_{\text{MSE}}$ was positively correlated with the number of species present per location (\cref{fig:mse_improvements}c, Pearson's $r=0.34$, $p=2 \cdot 10^{-6}$), showing that the CNN model's advantage over mean rates grows as number of species increases (but note that the model does not know the possible species \textit{a priori}). This is probably because locations with many species will inherently include less common species, such as habitat specialists, which the mean rate model poorly predicts. 

\begin{table*}[ht]
\begin{tabular}{lllll}
\toprule
Model & Top-10 [\%] & Top-5 [\%] & MSE [1e-02] \\
\midrule
Mean rate & 67.3 & 58.7 & 1.39 \\
Unfrozen 3-layer SeCo & 68.2 $\pm$ 0.2 & 61.8 $\pm$ 0.8 & 1.32 $\pm$ 0.03 \\
Frozen 3-layer SeCo & 69.6 $\pm$ 0.2 & 62.4 $\pm$ 1.0 & 1.21 $\pm$ 0.04 \\
Frozen 3-layer SeCo + CR (best grid) & 70.0 $\pm$ 0.2 & 62.3 $\pm$ 0.6 & 1.21 $\pm$ 0.03 \\
Frozen 3-layer SeCo + CR (best random) & \textbf{70.4 $\pm$ 0.2} & \textbf{63.0 $\pm$ 0.9} & \textbf{1.20 $\pm$ 0.03}\\
\bottomrule
\end{tabular}
\caption{Mean and SEM of validation metrics for different models with and without contrastive regularisation (CR). The best performing model for each metric is highlighted in bold.
\label{tab:model_changes}}
\end{table*}

\section{Discussion}
\subsection{Conclusion}
We set out to test whether 1) we could predict butterfly presence from satellite data and 2) contrastive regularisation would improve performance in this small-data regime. 
We found that a Resnet-based model was able to predict butterflies better than a mean rate model, and that contrastive regularisation further significantly improved the top-10 accuracy. 
However, as the best model reached 70.4\% top-10 accuracy and 63.0\% top-5 accuracy, there is scope for further improvements. 
To this end we make our new data set S2-BMS available to the community, to encourage further advances in this area.

\subsection{Data set}
We have created a new data set, S2-BMS, of Sentinel-2 satellite images and UKBMS butterfly encounter rates. Similar data sets have been created previously for bird and butterfly records in the USA and Kenya \citep{teng2023satbird, abdelwahed2024predicting} and plant records in the USA and France \citep{cole2020geolifeclef}, and S2-BMS complements these by considering new species, a new region and by using data collected according to very structured protocols \citep{UKBMS2018, UKBMS2019}. 
Specifically, UKBMS data are obtained from field surveys by trained volunteers using standardised methods to count butterflies along transects walked weekly during the summer in repeated years \citep{UKBMS2018, UKBMS2019}. The presence of a species at a site on any date depends on the species abundance as well as its detection likelihood, however given the expertise of the volunteers, we expect a relatively high probability of butterfly species being detected if they occur \citep{pocock2015biological}. However, seasonality and hence temporal abundance varies across species (for example
some butterfly species occur in flight (as adults) for a few weeks, whereas others are present for much longer through the season \citep{dennis2013indexing}). Another challenge is that the large area of interest per location (4 km$^2$) can lead to `noisy’ satellite images, depending on the size of relevant habitat per image.

\subsection{Contrastive learning}
Contrastive learning is predominantly used as a self-supervised pre-training task by augmenting the input data (images) \citep{chen2020simple}, but if labels are available, supervised contrastive learning can improve performance \citep{khosla2020supervised}. Remote sensing provides a vast source of unlabelled data, and as such contrastive learning has typically been used in a self-supervised setting \citep{manas2021seasonal, wang2022ssl4eo, jean2019tile2vec}. 
However, for biodiversity research, large quantities of geo-referenced citizen-science data are publicly available \citep{gbif_general}, collected by volunteers who log biodiversity records such as wildlife sightings \citep{sullivan2009ebird, UKBMS2018} and plants \citep{cole2020geolifeclef}. 
Hence, it is possible to create large, labelled remote sensing data sets in the biodiversity domain without additional annotation. 
This motivated us to develop a new supervised contrastive loss, which could further benefit the prediction accuracy of other taxonomic groups \citep{teng2023satbird, cole2020geolifeclef}.

\subsection{Model performance}
The best CNN model reached 70.4\% top-10 accuracy and 63.0\% top-5 accuracy on average (\cref{tab:model_changes}), and it performed better than the mean rate baseline for all species with mean encounter rates $> 0.01$ (\cref{fig:mse_improvements}b). It is unsurprising that the rarest species are the most difficult to predict, both given the strong class imbalance and also the limited size of the data set. However, for applications in conservation this might not be a large issue, as monitoring of rare species tends to be more complete than of (semi-)common species because of additional, species-specific monitoring programmes \citep{UKBMS2018}. 
Our accuracy metrics (compared to mean rate) are similar to the Resnet-18 RGB+NIR results on SatBird \citep{teng2023satbird}, which has more species but also more data points. There, model performance was further improved by including additional geospatial layers of bioclimatic and pedologic data, as well as range maps \citep{teng2023satbird}. 
We experimented with including contrastive regularisation, for which we derived PECL, a supervised contrastive loss for data with probability vector labels. We found that performance indeed moderately improved (most notably for top-10 accuracy, \cref{tab:model_changes}). 
Other future directions for improving species-presence prediction models could include transfer learning across data sets \citep{pan2009survey} (\textit{e.g.}, SatBird to S2-BMS), including location embeddings \citep{klemmer2023satclip} or using other pre-trained geospatial foundation models \citep{rolf2021generalizable, tseng2023lightweight, russwurm2024meta}, and closer interaction with species distribution models that reduce data bias \citep{dennis2013indexing, clarke2020new, fink2023double, bowler2022temporal, vanderplas2025ese}.

\subsection*{Data and software availability}
We have published the S2BMS data set on Zenodo: \href{https://doi.org/10.5281/zenodo.15198884}{doi.org/10.5281/zenodo.15198884}. All code is available at: \href{https://github.com/vdplasthijs/PECL/}{github.com/vdplasthijs/PECL/}. The original Sentinel-2 satellite data is publicly available on Google Earth Engine \citep{sentinel2data}. UKBMS data is publicly available on GBIF \citep{gbiforg_user_ukbms_2024}. 
\subsection*{Acknowledgements}
The authors thank Matthew Fry (UK Centre for Ecology and Hydrology) and Richard Fox (Butterfly Conservation) for constructive discussions, and thank the reviewers for their helpful suggestions. This work was funded by The Alan Turing Institute and the Engineering and Physical Sciences Research Council [grant number EP/Y028880/1]. TLvdP was partially funded by NWO grant \#184.036.014 for LTER-LIFE. 
{
    \small
    \bibliographystyle{ieeenat_fullname}
    \bibliography{main}
}

\end{document}